\documentclass[conference]{IEEEtran}
\IEEEoverridecommandlockouts
\usepackage{cite}
\usepackage{amsmath,amssymb,amsfonts}
\usepackage{algorithmic}
\usepackage{graphicx}
\usepackage{textcomp}

 \usepackage{multirow}
 \usepackage[table,xcdraw]{xcolor}

\usepackage{subcaption}
 \usepackage{dblfloatfix}

\usepackage{bigstrut}
\usepackage{multirow}
\usepackage{color}
\usepackage{caption}
\usepackage{fancyhdr}
\usepackage{listings} 
\usepackage{changepage}
\usepackage{pgfplots}
\usepackage{balance}
\usepackage{enumitem}
\usepackage{breqn}
\def\BibTeX{{\rm B\kern-.05em{\sc i\kern-.025em b}\kern-.08em
    T\kern-.1667em\lower.7ex\hbox{E}\kern-.125emX}}

\begin{document}
\title{Differential Newborn Face Morphing Attack Detection using Wavelet Scatter Network  \\
}

%

\author{Raghavendra Ramachandra\textsuperscript{1} \quad \quad  Sushma Venkatesh\textsuperscript{2} \quad \quad   Guoqiang Li \textsuperscript{3}  \quad \quad Kiran Raja \textsuperscript{1}\\
\textsuperscript{1}Norwegian University of Science and Technology (NTNU), Norway.
\textsuperscript{2}AiBA AS, Norway. \\
\textsuperscript{3}MOBAI AS, Norway.\\
{\tt\small email: \{raghavendra.ramachandra, kiran.raja\}@ntnu.no; sushma@aiba.ai; gl@mobai.bio}
}

\maketitle

\begin{abstract}
Face Recognition System (FRS) are shown to be vulnerable to morphed images of newborns. Detecting morphing attacks stemming from face images of newborn is important to avoid unwanted consequences, both for security and society. In this paper, we present a new reference-based/Differential Morphing Attack Detection (MAD) method to detect newborn morphing images using Wavelet Scattering Network (WSN). We propose a two-layer WSN with  250 $\times$ 250 pixels and six rotations of wavelets per layer, resulting in 577 paths. The proposed approach is validated on a dataset of 852 bona fide images and 2460 morphing images constructed using face images of 42 unique newborns. The obtained results indicate a gain of over 10\% in detection accuracy over other existing D-MAD techniques. 
\end{abstract}

\section{Introduction}
Morphing attacks using face images have demonstrated high degree of threat to automatic Face Recognition Systems (FRS). Morphing is the process of blending two or more face images, resulting in a constituent face image that visually resembles source face images used for morphing. The widespread availability of open source tools for morphing \cite{Venkatesh-MADSurvey-IEEETTS-2021, GMAP, ramachandra2022mask} further amplifies the threat, as it can facilitate morphing attack generation without the need for expert knowledge. It is well demonstrated in  \cite{godage2022analyzing} that morphed images can successfully deceive human observers, including trained experts like border guards and ID experts. Threats on FRS coupled with weakness of human observers in detecting morphed images intensify the attack strength on real-life applications including border control and remote ID verification. Morphing detection of newborn faces therefore becomes critical to avoid various problems such as illegal adoption, sexual exploitation, child marriages, and organ harvesting.

The severity of the problem has led to the development of Morphing Attack Detection (MAD) algorithms. MAD algorithms can be broadly classified into two types: Single image-based MAD (S-MAD) and Differential or reference-based MAD (D-MAD) \cite{Venkatesh-MADSurvey-IEEETTS-2021}. In S-MAD, an algorithm makes the decision using single image, whereas in D-MAD, the algorithm makes a decision using two images where one of the image is captured in a trusted environment (e.g., captured from Automatic Border Control (ABC) gate) and the second image being suspicious. While both algorithms have their own use cases in real-life applications, D-MAD techniques are more reliable given at-least one image is captured in trusted environment.

D-MAD algorithms have been extensively studied in the literature and can be broadly categorized into three types: (i) texture-based approaches, (ii) face demorphing, and (iii) deep learning feature based approaches. Texture-based methods are based on LBP \cite{Scherhag-FaceMorphingAttacks-TIFS-2020}, BSIF \cite{Scherhag-FaceMorphingAttacks-TIFS-2020},  differences in Facial-Landmarks \cite{Scherhag-LandmarkMAD-ICISP-2018}, scale-space features \cite{ramachandra2022residual}, and 3D facial textures \cite{Jag_ABC_gate_2019}. Facial demorphing techniques \cite{Ferrara-Demorphing-TIFS-2018} reverse the morphing process by using a reference image. Deep learning based approaches use of pre-trained deep Convolutional Neural Networks (CNN) \cite{JagIJCB2022, Scherhag-FaceMorphingAttacks-TIFS-2020},   Single and Double Siamese Networks \cite{Borghi-DoubleSiamese-mdpi-2021}  \cite{soleymani-DMAD-Siamese-2021}, Attention based networks \cite{Aghdaie-wavelet-subbandMAD-detection-2021}  GAN \cite{Peng-FD-GAN-IEEE-2019}, \cite{Banerjee-conditional-GAN-IJCB-2021} and  Auto-Encoders \cite{Aghdaie-wavelet-subbandMAD-detection-2021}.  Readers can refer to a recent survey for a detailed overview of D-MAD techniques \cite{Venkatesh-MADSurvey-IEEETTS-2021}.

\begin{figure}[t!]
\begin{center}
\includegraphics[width=1.0\linewidth]{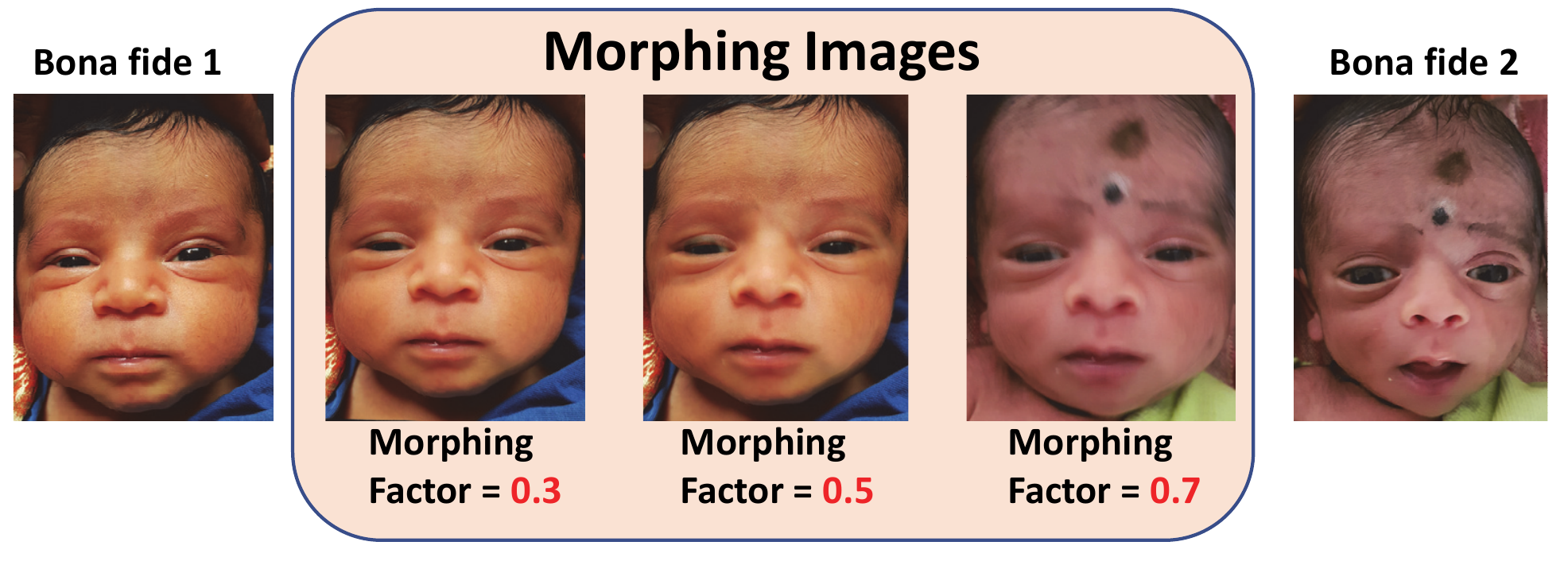}
\end{center}
   \caption{Illustration of newborn face morphing with different morphing factors}
\label{fig:Intro}
\end{figure}

Even though D-MAD techniques are widely studied in the literature, all these techniques are limited to normal (or adult) face morphing detection.  To the best of our knowledge, there have been no works reporting D-MAD approaches for face images of newborns. Unlike adult face images, morphing detection for newborn face images need to address the various challenges such as pose, expression, external marks on face and lesser identity features making it a challenge. Figure \ref{fig:Intro} illustrates example morphed images of a newborn faces with three different morphing factors. Early work on newborn face morphing \cite{infantMorphing2021} demonstrated the vulnerabilities in FRS. The baseline performance  of the S-MAD techniques were also presented in the same work which indicated a degraded performance in detection as compared to the MAD for normal (or adult) faces. We are therefore motivated to address this problem by introducing a new D-MAD method for detecting morphed images of newborn faces. 

We propose a new D-MAD algorithm based on features extracted using a Wavelet Scattering Network (WSN) in this work. The WSN features are both invariant to scale and translation and thus assert it suitable for newborn face D-MAD which often have challenges in pose and expression. We validate our assertion with an experimental analysis using a morphed face image dataset of newborns. the main contributions of this work is as follows: 
\begin{itemize} [leftmargin=*,noitemsep, topsep=0pt,parsep=0pt,partopsep=0pt]
\item Novel method for the differential morphing attack detection tailored to the new born identities. The proposed method is based on the Wavelet Scattering Network (WSN), which can extract time- and scale-invariant features from the color space representation of the newborn face image. 
\item Extensive experimental validation of proposed approach on the infant face dataset \cite{infantMorphing2021} consisting of 852 bona fide capture and 2460 morphing samples obtained from 42 unique identities. Morphing was performed at three different morphing factors 0.3, 0.5 and 0.7. 
\item The detection performance of the proposed method is benchmarked with the deep facial features \cite{Scherhag-FaceMorphingAttacks-TIFS-2020} and obtained results indicate better performance with a gain over 10\% in Detection-Equal Error Rate (D-EER). 
\end{itemize}

The rest of the paper is organised as follows: Section \ref{sec:prop} presents the proposed D-MAD technique on newborn face identities, Section \ref{sec:exp} discuss the experimental results and Section \ref{sec:conc} draws the conclusion.

\section{Proposed Method}
\label{sec:prop}

\begin{figure}[htp]
\begin{center}
\includegraphics[width=0.9\linewidth]{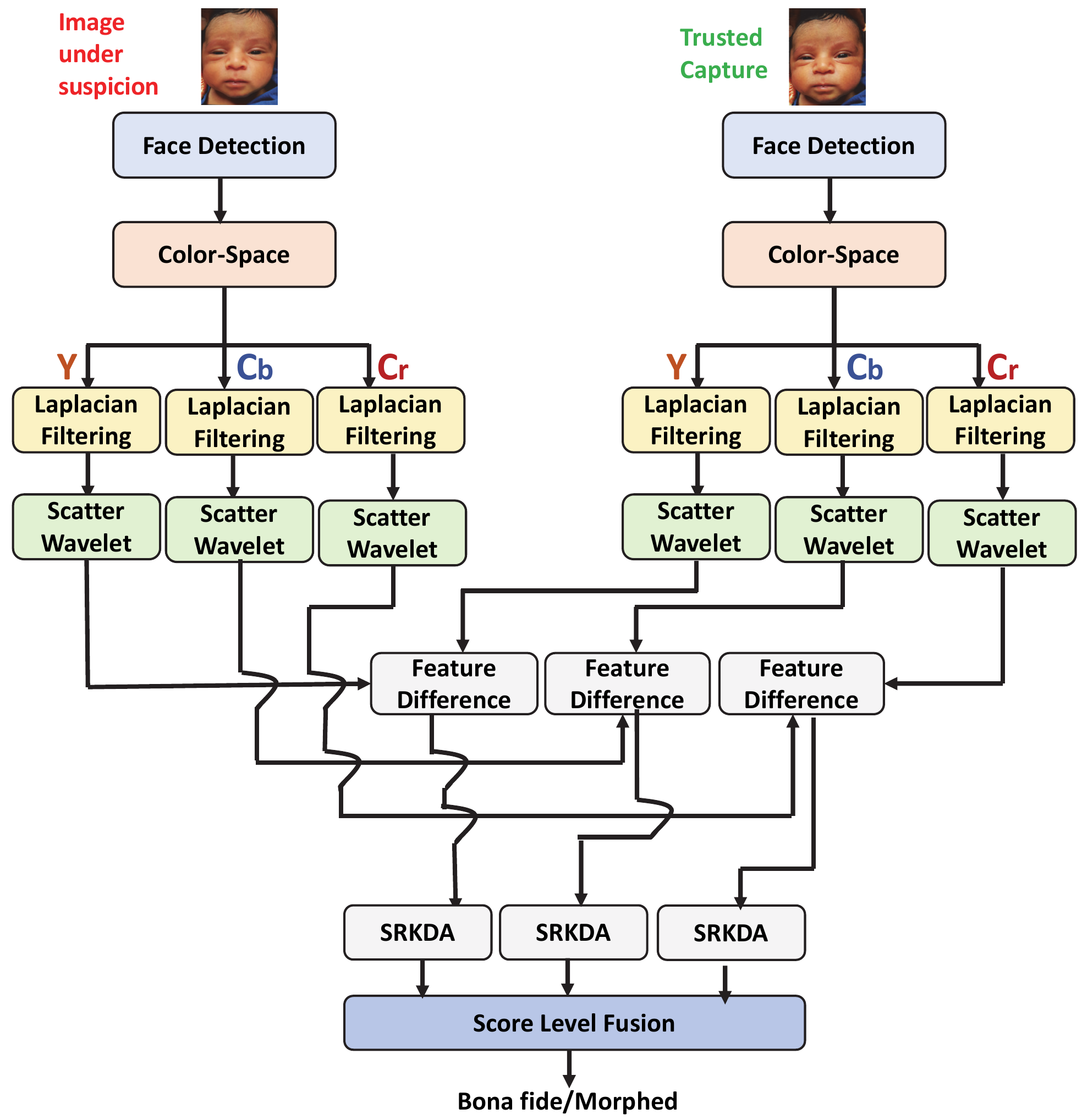}
\end{center}
   \caption{Block diagram of the proposed D-MAD algorithm for newborn face images}
\label{fig:Prop}
\end{figure}

\begin{figure}[htp]
\begin{center}
\includegraphics[width=0.9\linewidth]{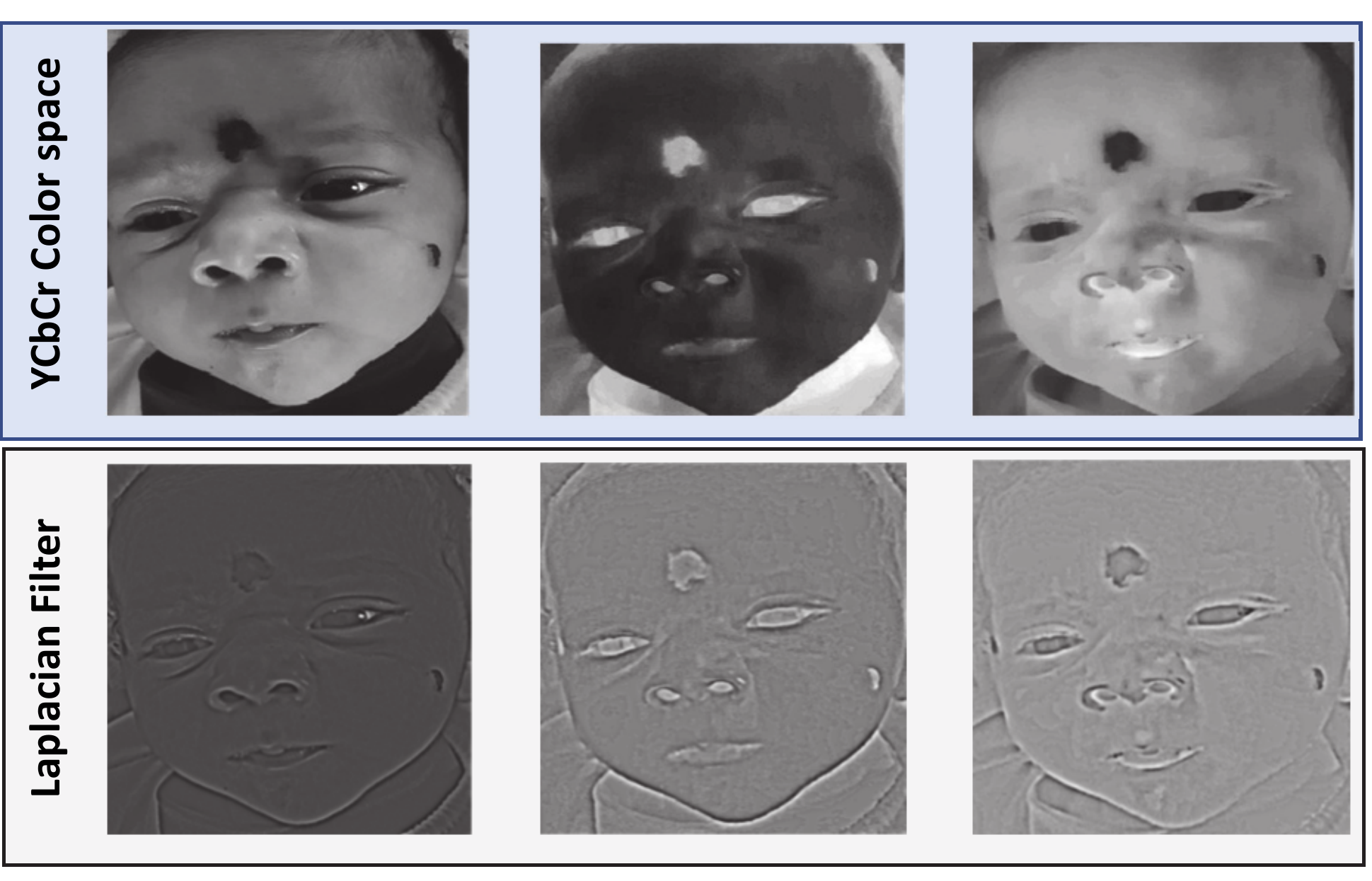}
\end{center}
   \caption{Qualitative results of the proposed method with color space and Laplacian filtering}
\label{fig:Quali}
\end{figure}

Figure \ref{fig:Prop} shows the block diagram of the proposed D-MAD technique for newborn identities. The proposed method can be structured in seven different functional block including face detection, color-space representation, scale-space representation using Laplacian transform, feature extraction using scatter wavelet, feature difference, classification using Spectral Regression Kernel Discriminant Analysis (SRKDA) and score level fusion. The proposed method considers two input images, reference image $I_{r}$ and trusted capture image $I_{t}$ which are processed to extract the features independently.

Given images $I_{s}$ and $I_{t}$ corresponding to suspicious image (morphed or bona fide) and trusted image respectively, face detection is carried out using MTCNN \cite{MTCNNPaper} by considering its robustness to pose and resolution. Face detection is performed independently on  $I_{s}$ and $I_{t}$ to obtain the corresponding face regions denoted as $F_{s}$ and $F_{t}$ respectively. In the next step, we obtain color-space representation of $F_{s}$ and $F_{t}$ independently. In this work, we choose $YC_{b}C_{r}$ by considering its application to morphing attack detection presented in earlier work \cite{ramachandra-algorithmic-fairness-CVPR-2022}. The use of  $YC_{b}C_{r}$ provides discriminate features that can highlight pixel discontinuities. Thus, the color space representation of $F_{s}$ and $F_{t}$ can be represented as $Y_{s}$, $B_{s}$, $R_{s}$ and $Y_{t}$, $B_{t}$, $R_{t}$ respectively for suspicious and trusted image. Figure \ref{fig:Quali} shows the qualitative results of the color space for the example of face image. In the next step, we process independent color channels to extract high-frequency features using Laplacian filtering \cite{raghavendra2022multimodality}. We employ Laplacian filtering as it can extract rich information on the edge discontinuities and localize double edges that are due to the morphing process. We perform the Laplacian filtering independently on color channels that will results in $L_{s}^{Y}$, $L_{s}^{B}$, $L_{s}^{R}$ and $L_{t}^{Y}$, $L_{t}^{B}$, $L_{t}^{R}$ respectively. Figure \ref{fig:Quali} shows the qualitative results of the Laplacian filtering results of the example face image.

\begin{table*}[htp]
\centering
\resizebox{.75\textwidth}{!}{%
\begin{tabular}{|l|l|l|ll|}
\hline
\rowcolor[HTML]{DAE8FC} 
\cellcolor[HTML]{DAE8FC} &
  \cellcolor[HTML]{DAE8FC} &
  \cellcolor[HTML]{DAE8FC} &
  \multicolumn{2}{l|}{\cellcolor[HTML]{DAE8FC}\textbf{BPCER @ APCER =}} \\ \cline{4-5} 
\rowcolor[HTML]{DAE8FC} 
\multirow{-2}{*}{\cellcolor[HTML]{DAE8FC}\textbf{Algorithms}} &
  \multirow{-2}{*}{\cellcolor[HTML]{DAE8FC}\textbf{Morphing Factor}} &
  \multirow{-2}{*}{\cellcolor[HTML]{DAE8FC}\textbf{D-EER (\%)}} &
  \multicolumn{1}{l|}{\cellcolor[HTML]{DAE8FC}\textbf{5\%}} &
  \textbf{10\%} \\ \hline
 &
  0.3 &
  30.26 &
  \multicolumn{1}{l|}{80.90} &
  67.98 \\ \cline{2-5} 
 &
  0.5 &
  34.52 &
  \multicolumn{1}{l|}{85.14} &
  76.14 \\ \cline{2-5} 
\multirow{-3}{*}{Deep Features \cite{Scherhag-FaceMorphingAttacks-TIFS-2020}} &
  0.7 &
  37.18 &
  \multicolumn{1}{l|}{85.88} &
  74.53 \\ \hline
 &
  0.3 &
  \cellcolor[HTML]{EFEFEF}\textbf{24.72} &
  \multicolumn{1}{l|}{\cellcolor[HTML]{EFEFEF}\textbf{83.57}} &
  \cellcolor[HTML]{EFEFEF}\textbf{24.72} \\ \cline{2-5} 
 &
  0.5 &
  \cellcolor[HTML]{EFEFEF}\textbf{22.85} &
  \multicolumn{1}{l|}{\cellcolor[HTML]{EFEFEF}\textbf{78.22}} &
  \cellcolor[HTML]{EFEFEF}\textbf{48.52} \\ \cline{2-5} 
\multirow{-3}{*}{Proposed Method} &
  0.7 &
  \cellcolor[HTML]{EFEFEF}\textbf{24.72} &
  \multicolumn{1}{l|}{\cellcolor[HTML]{EFEFEF}\textbf{82.63}} &
  \cellcolor[HTML]{EFEFEF}\textbf{61.43} \\ \hline
\end{tabular}%
}

\caption{Quantitative detection performance of the proposed approach compared to other existing D-MAD techniques on the newborn dataset}
\label{tab:results}
\end{table*}


In the next step, we extract the features using a Wavelet Scattering Network (WSN) \cite{ScatterWavelet} independently on each filtered image.  In this work, we construct a two-layer image scatter network with a $250 \times 250$ pixels with invariant scale such that two wavelets per octave in the first layer and one wavelet per octave in the second layer. Furthermore, we use six rotations of wavelets per layer. Therefore, the WSN used in this work has 577 paths. Let the WSN computed for Laplacian filtered images be denoted as: $W_{s}^{Y}$, $W_{s}^{B}$, $W_{s}^{R}$ and $W_{t}^{Y}$, $W_{t}^{B}$, $W_{t}^{R}$ respectively for suspicious and trusted image. In the next step, we compute the unsigned feature difference between the WSN features computed from the corresponding morphed and trusted image. Let the feature difference be denoted as $FD_{Y} = W_{s}^{Y} -  W_{t}^{Y}$, $FD_{Y} = W_{s}^{B} -  W_{t}^{B}$  and $FD_{Y} = W_{s}^{R} -  W_{t}^{R}$. The computed feature differences are then used to learn a Spectral Regression Kernel Discriminant (SRKDA) classifier \cite{SRKDAPaper} to differentiate morphed and trusted image. The classifier provides a comparison scores corresponding to the three different feature differences be $S_{1}, S_{2} $  and $S_{3}$ respectively. Finally, we fuse the scores using the sum-rule, i.e., $F_{S} =  \sum_{t = 1}^{3} S_{i}$ to obtain the final score. The final score is then used to decide whether a given image under question is morphed or bona fide.

\section{Experiments and Results}
\label{sec:exp}

In this section, we discuss the quantitative results of the newborn morphing detection especially in the reference-based scenario. Experiments are performed using the newborn face dataset \cite{infantMorphing2021} comprised of 42 unique data subjects captured in multiple sessions. The performance of the  D-MAD techniques are presented using the
ISO/IEC 30107- 3 metrics \cite{ISO-IEC-30107-3-PAD-metrics-170227} such as Attack Presentation Classification Error Rate (APCER) and Bona fide Presentation Classification Error Rate (BPCER). APCER is defined as the proportion of attack presentations incorrectly classified as bona fide, whereas BPCER is defined as the proportion of the bona fide incorrectly classified as attack presentation. The D-EER indicates the value at which proportion of APCER equals the proportion of BPCER. The detection performance of the proposed method is benchmarked with the existing method based on deep face features \cite{Scherhag-FaceMorphingAttacks-TIFS-2020}. We particularly consider deep face features \cite{Scherhag-FaceMorphingAttacks-TIFS-2020} as it demonstrated the best detection performance in NIST benchmark under D-MAD category\cite{Nist-Frvt-Morph}.

To evaluate the detection performance of the proposed and existing methods effectively, the entire dataset was divided into two disjoint sets. The training set consisted of  20 unique data subjects and the testing set consisted of 22 unique data subjects. In this work, the face morphing process was carried out using a landmark-based face morphing tool \cite{Ferrara-TextureBlendingAndShapeWarpingInFaceMorphing-IEEE-BIOSIG-2019} by considering its ability to generate high-quality morphing images, resulting in high vulnerability across different FRS \cite{zhang-MIPGAN-TBIOM-2021}. Morphing images were generated using three different morphing factors: 0.3, 0.5, and 0.7. Thus, the training dataset consisted of 310 bona fides and 367 $\times$ 3 (with different morphing factors) = 1101 morphing images. The testing dataset consisted of 542 bona fides and 453 $\times$ 3 (with different morphing factors) = 1359 morphing images. Thus, the entire dataset consisted of 852 bona fide and 2460 morphing images corresponding to 42 unique newborn identities.

\begin{figure}[htp]
\minipage{0.4\textwidth}
  \includegraphics[width=0.85\linewidth]{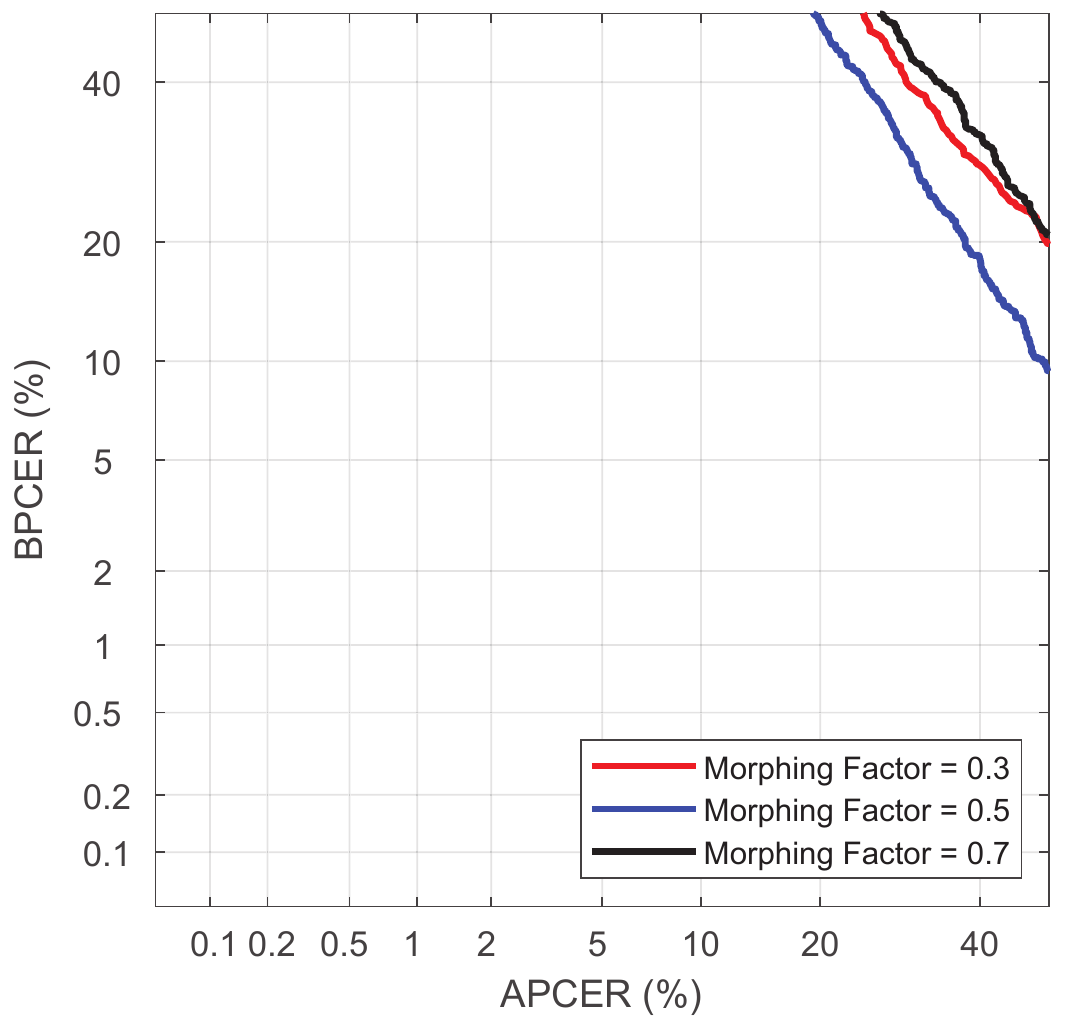}
  \caption{Detection Error Trade-off (DET) curves depicting performance of the
D-MAD approach based on Deep Features \cite{Scherhag-FaceMorphingAttacks-TIFS-2020}}\label{fig:Deep}
\endminipage
\hfill
\minipage{0.4\textwidth}
  \includegraphics[width=0.85\linewidth]{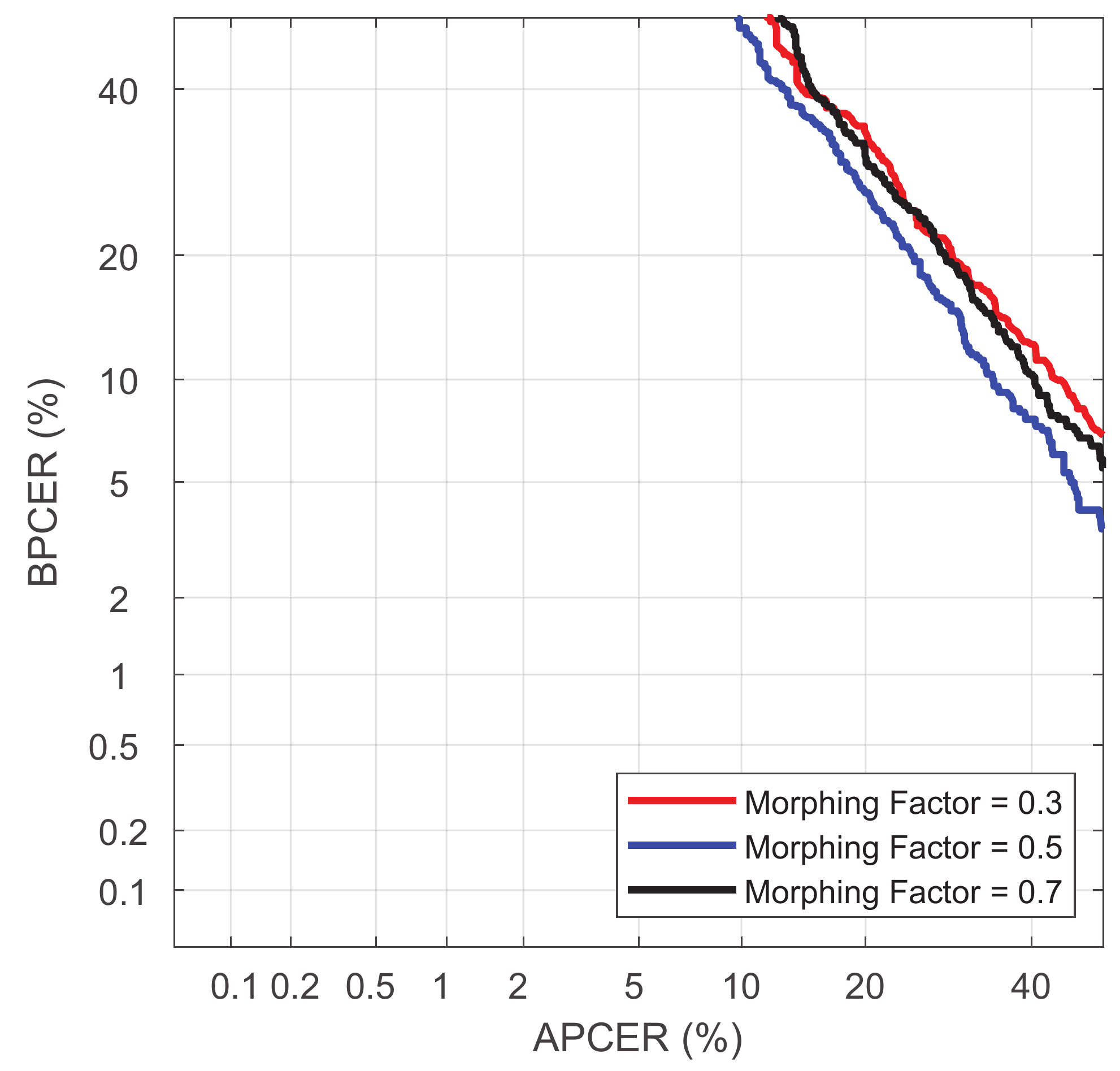}
  \caption{Detection Error Trade-off (DET) curves depicting performance of the proposed D-MAD approach.}\label{fig:Proposed}
\endminipage
\end{figure}

Table \ref{tab:results} shows the quantitative performance of the proposed method and existing method for newborn morphing attack detection. Figure \ref{fig:Deep} and \ref{fig:Proposed} show the DET curves for the existing method (deep face features \cite{Scherhag-FaceMorphingAttacks-TIFS-2020}) and the proposed method with  different morphing factors.  Based on the obtained results following are the important observations: 
\begin{itemize} [leftmargin=*,noitemsep, topsep=0pt,parsep=0pt,partopsep=0pt]
\item The morphing detection performance of the proposed and existing methods vary with morphing factors. The variation in performance across different morphing factors for the proposed method indicates less variation compared with the existing method.  The higher variation in the detection performance of the existing method can be attributed to the lack of identity features suitable for detecting morphing attacks as the existing method is based on deep face features extracted using ArcFace FRS \cite{Scherhag-FaceMorphingAttacks-TIFS-2020}. 
\item Among the three different morphing factors, the proposed method obtains the lowest D-EER (\%) when morphing factor is $0.5$. However, the existing method indicated the lowest D-EER (\%) when the morphing factor was $0.3$. A morphing factor is considered to generate realistic attack to deceive human observers. As it can be observed, the proposed approach while obtaining better D-EER than existing approach can still achieve better performance in the realistic case of morphing factor of $0.5$. 
\item The proposed method shows the best performance compared with the existing method on all three morphing factors. The best performance of the proposed method was noted with a morphing factor of $0.5$, with D-EER = 22.85\%. The proposed approach can be seen to gain an average of  10\% detection accuracy over the compared method indicating the superiority. 
\end{itemize}

\section{Conclusion}
\label{sec:conc}
Reliable detection of morphing attacks on newborns is challenging because of the lack of identity features. In this study, we present a novel method based on Wavelet Scattering Network (WSN) that can extract time and scale-invariant features. The proposed D-MAD approach takes two facial images and processes to extract the color space using YCbCr. The independent color channel image is further processed using Laplacian filters to extract edge discontinuities that result from morphing process. A 2-layer WSN is employed to extract the discriminant features and the computed feature difference between two facial images is used to learn a SRKDA classifier.  The comparison scores on the individual color channels are further combined using the sum rule to obtain the final score. Extensive experiments conducted on the newborn morphing dataset indicate an improved detection performance of the proposed method with an average gain over 10\% over compared state-of-the-art method.

{\small
\bibliographystyle{IEEEtran}
\bibliography{Face_Morph_references}
}

\end{document}